\documentclass[11pt,a4paper]{article}
\usepackage[hyperref]{acl2018}
\usepackage{times}
\usepackage{latexsym}

\usepackage{url}

\usepackage{amssymb}
\usepackage{amsmath}
\usepackage{graphicx}

\aclfinalcopy 

\title{Abstractive Summarization Using Attentive Neural Techniques}

\author{Jacob Krantz \\
  Gonzaga University \\
  Spokane, WA \\
  {\tt jkrantz@zagmail.gonzaga.edu} \\\And
  Jugal Kalita \\
  University of Colorado, Colorado Springs\\
  Colorado Springs, CO \\
  {\tt jkalita@uccs.edu} \\}

\date{}

\begin{document}
\maketitle
\begin{abstract}
In a world of proliferating data, the ability to rapidly summarize text is growing in importance. Automatic summarization of text can be thought of as a sequence to sequence problem. Another area of natural language processing that solves a sequence to sequence problem is machine translation, which is rapidly evolving due to the development of attention-based encoder-decoder networks. This work applies these modern techniques to abstractive summarization. We perform analysis on various attention mechanisms for summarization with the goal of developing an approach and architecture aimed at improving the state of the art. In particular, we modify and optimize a translation model with self-attention for generating abstractive sentence summaries. The effectiveness of this base model along with attention variants is compared and analyzed in the context of standardized evaluation sets and test metrics. However, we show that these metrics are limited in their ability to effectively score abstractive summaries, and propose a new approach based on the intuition that an abstractive model requires an abstractive evaluation.
\end{abstract}

\section{Introduction}
The goal of summarization is to take a textual document and distill it into a more concise form while preserving the most important information and meaning. To this end, two approaches have historically been taken; extractive and abstractive. Extractive summarization selects the most important words of a given document and combines and rearranges them to form a final summarization \cite{nallapati2017summarunner}. This approach is restricted to using words directly from the source document and so is unable to paraphrase. Abstractive algorithms generate a summary from an attempt to understand a document's meaning, allowing for paraphrasing much like a human may do. Abstractive approaches are more difficult to develop than extractive ones because an intermediate representation of knowledge is required. As such, dominant techniques of summarization have been extractive in nature, with wide-ranging solutions utilizing statistical, topic-based, graph-based, and machine learning approaches \cite{gambhir2017recent}. With the potential for generating more coherent and insightful summaries, abstractive approaches are gaining in popularity fueled by novel deep learning techniques \cite{see2017get}. The abstractive summarization process includes converting words to their respective embeddings, computing a document representation, and generating output words. Neural networks have recently been shown to perform well for every step \cite{dong2018survey}.

In deep learning models, attention allows a decoder to focus on different segments of an input while stepping through output regions. In the related sequence to sequence task of machine translation, attention was introduced to the existing encoder-decoder model \cite{bahdanau2014neural}. This resulted in large improvements over past systems due to the ability to consider a larger window of context during the output generation. Progressing this further, \citet{vaswani2017attention} showed that multi-headed self-attention can replace recurrence and convolutions entirely. As the areas of machine translation and abstractive summarization are related both structurally and semantically, the developments in machine translation may inform the direction of research in abstractive summarization. In this paper, we apply these advancements and develop them further in pursuit of sentence summarization. In any attempt at summarization, the resulting text must be much more condensed than the original. In this task, all generated summaries are constrained to a fixed maximum length so that tested models must learn how to decide what information should be reproduced.

\section{Related Work}
Successful sentence summarization approaches have classically used statistical methods. TOPIARY \cite{zajic2004bbn} detected salient topics that guided sentence compression while using linguistic transformations. MOSES, a statistical machine translation system, also performed well when directly used for summarization \cite{koehn2007moses}. Attention mechanisms have been shown to improve the results of abstractive summarization. \citet{rush2015neural} improved over classic statistical results by using a neural language model with a minimal contextual attention encoder. After the primary model training, an extractive tuning step was performed on an adjacent dataset. A related extension of this used a convolutional attentive encoder and experimented with replacing the decoder language model with RNN variants. LSTM cells and RNN-Elman both showed improved ROUGE scores \cite{chopra2016abstractive}. An attentive encoder-decoder was also employed by \citet{zeng2016efficient} with one RNN architecture to re-weight another to improve context across the input sequence. Their decoder used attention with a copy mechanism that differentiated between out of vocabulary words based on their usage in the input. \citet{nallapati2016abstractive} continued progress on encoder-decoder architectures by employing a bidirectional GRU-RNN encoder with a unidirectional GRU-RNN decoder. Imposing dynamic vocabulary restrictions also improved results while reducing the dimensionality of the softmax output layer. Pointer-generator networks encode with a bidirectional LSTM and decode with attention restriction. A coverage vector that limits the attention of words previously attended over is maintained \cite{see2017get}.

Recently, summarization has made progress at the paragraph level due to reinforcement learning. A recurrent abstractive summarization model used teacher forcing and a similarity metric that compared the generated summary with the target summary \cite{paulus2017deep}. The architecture contained a bi-directional LSTM with intra-attention. Actor-critic reinforcement learning was used by \citet{li2018actor} to produce the highest scores for sentence summarization. One important consideration when optimizing purely on the test metric is that while overall recall is improved, higher ROUGE scores do not necessarily correlate with the readability of summaries.

\section{Models}
Encoder-decoder architectures provide an adaptable structure for the development of systems that solve sequence to sequence problems. The encoder maps the input sequence to a latent vector representation. The decoder takes this representation, called the context vector, and generates the output sequence. The models and their variants that follow are structured as such. We select a base architecture that provides a strong foundation on which to analyze the effect of self-attention variants. 

\begin{figure}[t]
  \centering
  \includegraphics[scale=0.41]{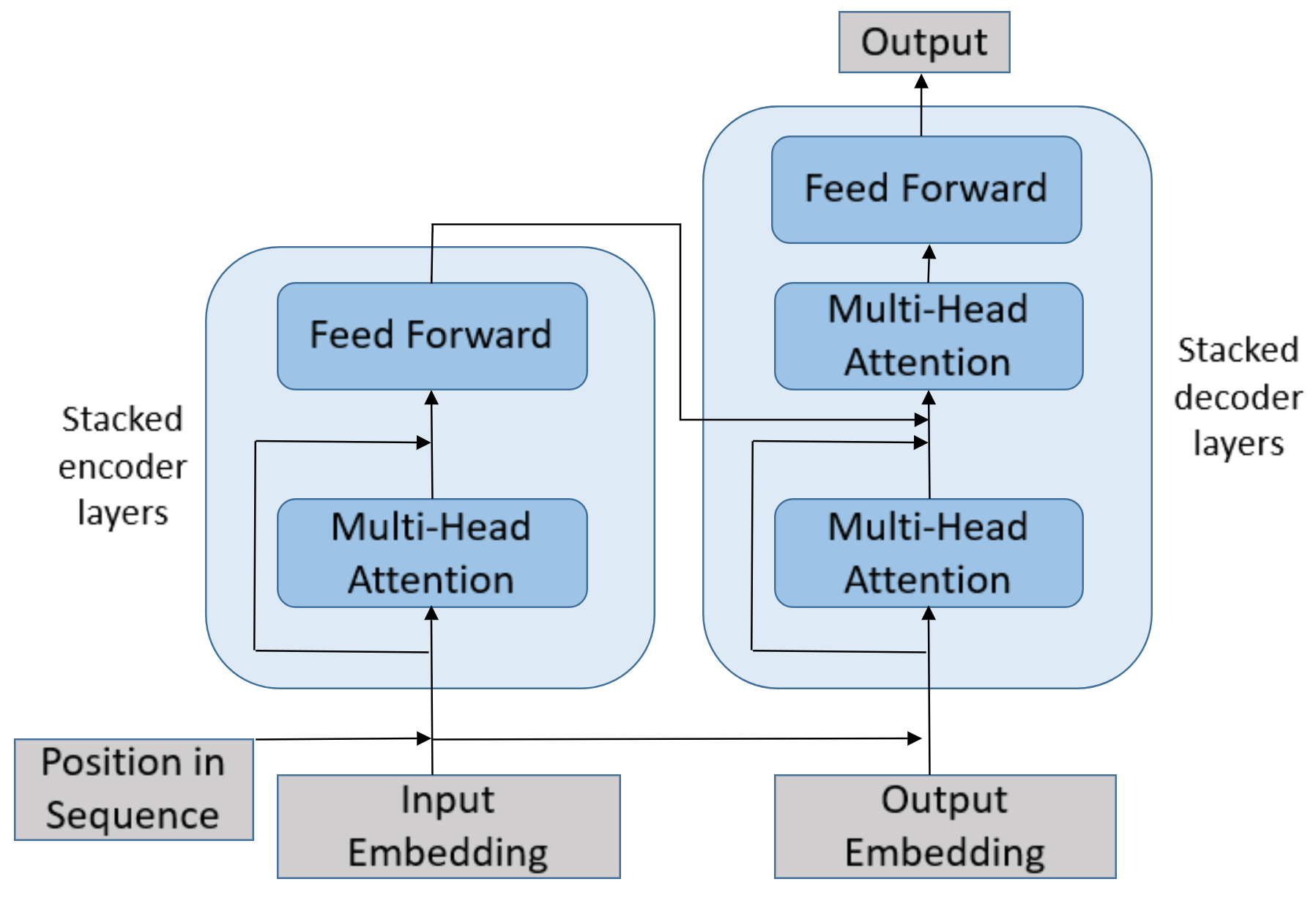}
  \caption{Transformer-based network architecture. The multi-headed attention mechanisms contain various recall options similar to and that expand upon \citet{vaswani2017attention}.}
\end{figure}

\subsection{The Transformer}
The Transformer architecture as proposed by \citet{vaswani2017attention} is notable for performing state of the art Machine Translation, and is more efficient to train than past systems by orders of magnitude. This is made possible by replacing sequence aligned recurrence with parallel self-attention. The sequence order is preserved in the self-attention modules by including positional embeddings. Instead of incremental values, the positional embeddings are determined by position on a sinusoidal time series curve. Further, masking of the decoder self-attention is performed, making the output of the next token dependent on that which has already been generated. Multi-headed self-attention is used in both the encoder and decoder. These mechanisms map a query vector to a key-value vector pair which results in an output vector. Tying together the encoder and decoder is a third multi-headed attention mechanism. The query comes from the self-attentive output of the decoder, and the keys and values from the self-attentive output of the encoder. In the work done by \citet{vaswani2017attention}, all attention heads used scaled dot-product attention, which is computationally efficient as multiple query, key, and value vectors can be implemented as a combined matrix. Scaled dot-product attention also defines the structure for the self-attention mechanisms we present below.

\begin{equation}
	attention = softmax(\frac{QK^{T}}{\sqrt{d}})V
\end{equation}

Many other attention mechanisms exist beyond the base dot-product attention. We analyze the performance of these mechanisms in the context of abstractive summarization. Changing the way the query, key, and value vectors interact allows an attention mechanism to learn different relationships between sequence elements.

\textit{Relative dot-product attention} uses scaled dot-product attention, but instead of using absolute positional encodings, uses a relative positional encoding. These relative encodings learn to relate the elements of the query to both the elements of the keys and values \cite{gehring2017convolutional}. The encodings can be distance-limited to a context window in the vector sequences.  

\textit{Local attention} divides the key-value vectors into localized blocks \cite{liu2018generating}. Each query is strided over a corresponding block with a given filter size. Blocks can contain positions both prior to and following a given position, thereby not masking any element based on absolute position. Self-attention is performed over each block in isolation. 

\textit{Local masked attention} adds a mask to the blocks of local attention. Blocks in a future sequential position are masked from the query but all elements within a block remain visible to a given query position. Intuitively, masking future positions forces a mechanism to attend to current and past positions, which may be an important restriction of the attention distribution.

\textit{Local block masked attention} masks both previous blocks and future blocks for a query position. Further, future positions within individual blocks are masked.

\textit{Dilated attention} also divides the key-value vectors into blocks, but introduces a gap in between each block. Each query position is limited to a context window of a specified number of blocks both preceding and following the memory position.

\textit{Dilated masked attention} performs the same operations as dilated attention and masks future memory positions within each block.

\begin{table*}[t]
  \centering
  \begin{tabular}{|c|l|l}
  \hline
  \textit{Target} & \textit{Endeavour astronauts join two segments of International Space Station.} \\
  Gen1   & Endeavour astronauts join two \textbf{sections} of International Space Station. \\ 
  Gen2   & Endeavour astronauts \textbf{remove} two segments of International Space Station. \\ 
  Gen3   & Endeavour astronauts join two segments of International Space Station. \\ \hline
  \end{tabular}
  \label{vert-sentences}
\end{table*}

\begin{table*}[t]
  \centering
  \begin{tabular}{|c|r|r|r|r|r|r|}
  \hline
  Sentence & ROUGE-1 & ROUGE-2 & ROUGE-l & Cos-Sim & WMD   & VERT   \\ \hline\hline
  Gen1     & 88.89   & 75.00   & 88.89   & 0.979   & 0.418 & 94.77  \\ 
  Gen2     & 88.89   & 75.00   & 88.89   & 0.924   & 0.512 & 91.08  \\ 
  Gen3     & 100.00  & 100.00  & 100.00  & 1.000   & 0.000 & 100.00 \\ \hline
  \end{tabular}
  \caption{Highlighted differences between ROUGE and VERT scoring. Notice that an incorrect word replacement (\textit{Gen2}) scores the same as a reasonable word replacement (\textit{Gen1}) in ROUGE. VERT discounts the score of \textit{Gen2} accordingly. \textit{Gen3} is included to show the perfect scores for an identical summary.}
  \label{table:vert-scores}
\end{table*}

\section{Evaluation}
The standard test metric for automatic summary generation is ROUGE, or Recall-Oriented Understudy for Gisting Evaluation \cite{lin2004rouge}. Before the ROUGE metrics were introduced, human judges were used for summary evaluation. Human judges provide an ideal evaluation, but are impractical for regular use. ROUGE allows for automatic comparison of generated summaries to target summaries, where target summaries are human-generated. Limited-length recall is commonly reported using ROUGE-1, ROUGE-2, and ROUGE-L. ROUGE-1 and ROUGE-2 compare unigram and bigram overlap, respectively. This generalizes to ROUGE-N for n-gram overlap. ROUGE-L determines the longest common subsequence (LCS). Evaluation quality of summarization models can be directly compared to previous work because the same metrics were reported for past models by \citet{rush2015neural}, \citet{zeng2016efficient}, \citet{nallapati2016abstractive}, \citet{li2018actor}, and others. These metrics allow for reasonably accurate comparison of summary generation models, but inherent problems exist. One critical limitation is that ROUGE does not consider equivalent paraphrasing or synonymous concepts. Since ROUGE works at the word level, meaning can only be captured and compared in a binary manner; either a word appears in the generated summary or it does not.

ROUGE 2.0 was proposed to alleviate this problem as well as remove the expectation that generated summaries need to be identical to the target summary \cite{ganesan2015rouge}. As pointed out by \citet{rush2015neural}, even the best human evaluator scored just 31.7 ROUGE-1 on the DUC2004 dataset. This illustrates the idea that two summaries do not need to be the same in order for both to be of high quality. Thus, a more appropriate approach to summary comparison may be to evaluate the semantic similarity between the generated and target summaries instead of using isolated word counts. ROUGE 2.0 captures semantic similarity using a synonym dictionary while still evaluating n-grams and LCS. While this addresses the word-level shortcoming of the original ROUGE metrics, similarity is still fixed to a discrete list of acceptable alternatives, which does not fully capture phrase substitution. A further improvement could be to evaluate the semantic similarity between two entities on a continuous scale.

\subsection{VERT Metric}
To improve the quality of summary evaluation, we introduce the VERT metric\footnote{Our VERT implementation is made publicly available at: \url{https://github.com/jacobkrantz/VertMetric}}, an evaluation tool that scores the quality of a generated hypothesis summary as compared to a reference target summary. VERT stands for Versatile Evaluation of Reduced Texts. VERT compares summaries on their underlying semantics rather than word count ratios. To calculate a VERT score for a summary pair, a similarity sub-score and dissimilarity sub-score are calculated and functionally combined. Naturally, a higher similarity score and a lower dissimilarity score leads to a higher, better VERT score. The similarity sub-score considers the semantics of each summary taken at the document level. A sentence embedding vector is synthesized for both generated and target summaries, and the cosine similarity between these two vectors provides the similarity score. The sentence embeddings are generated using \textit{InferSent}, an open-source neural encoder trained on natural language inference tasks \cite{conneau2017supervised}. \textit{InferSent} was chosen because it has been shown to generalize well for use in various problems requiring sentence representations. The dissimilarity sub-score operates at the individual word level rather than at the sentence level. An aggregate Euclidean distance is calculated between the words of the generated summary and the words of the target summary. This is done using the word mover's distance (WMD) algorithm, a measure of how far document A must travel to match document B within a word vector space \cite{kusner2015word}. Stop words are discarded prior to the distance calculation as their effect on the distance between documents is negligible. 

\subsection{Sub-Score Motivations}
A consideration would be to use just one of the two sub-scores as they are independent calculations. However, both the \textit{InferSent} cosine similarity and WMD are made more robust by the presence of the other score. WMD is unaffected by word ordering, whereas the encoder of \textit{InferSent} maintains sequential input. To illustrate, suppose the target sentence is \textit{``go right and then left''} and the generated sentence switches the order, stating \textit{``go left and then right.''} WMD  gives this a perfect distance of 0.0 but the \textit{InferSent} similarity more accurately discounts the score by 4.3\%. On the other hand, when longer summaries are compared, \textit{InferSent} embeddings begin to lose the effect of individual words because the word embeddings are replaced with a singular embedding. This is less of a problem for WMD. Finally, the similarity sub-score uses GloVe embeddings\footnote{\url{https://nlp.stanford.edu/projects/glove/}} pre-trained on Common Crawl while the dissimilarity sub-score uses Word2Vec\footnote{\url{https://code.google.com/archive/p/word2vec/}} trained on the Google News dataset. Using different word embeddings provides resistance to potential learned representation biases.

\subsection{Formula Specification}
The similarity sub-score is defined as $sim(s_{1},s_{2}) = cos(encode(s_{1}), encode(s_{2}))$ and the dissimilarity sub-score is defined as $dis(s_{1},s_{2}) = min(wmd(s_{1},s_{2}), \alpha)$. The maximum dissimilarity value $\alpha$ is the default distance when all of the generated words are out of vocabulary. Without this default, summaries with no words to compare would have an infinite distance and too strongly influence VERT score averages. Resulting sub-score values range as such: $0.0 \leq sim(s_{1},s_{2}) \in \mathbb{R} \leq 1.0$, and $0.0 \leq dis(s_{1},s_{2}) \in \mathbb{R} \leq \alpha$. We seek to combine these scores such that the final VERT score can be treated as a percentage: $0.0 \leq VERT(s_{1},s_{2}) \in \mathbb{R} \leq 1.0$. Further, $sim(s_{1},s_{2})$ and $dis(s_{1},s_{2})$ should be given equal weight in the final VERT score. To satisfy both criteria, we present the VERT equation:
\begin{equation}
  \begin{aligned}
    VER&T(s_{1},s_{2}) = \\
        &\frac{1}{2}(1+(sim(s_{1},s_{2})-\frac{1}{\alpha}dis(s_{1},s_{2})))
  \end{aligned}
\end{equation}
where $\alpha = 5.0$. The dissimilarity is normalized by $\alpha$ and the outer linearity, as multiplied by $\frac{1}{2}$, shifts the range from $[-1.0,1.0]$ to $[0.0,1.0]$. For the choice of $\alpha$, we observe an empirical distance ceiling of $5.0$ in Table \ref{table:wmd}. Incorporating this ceiling gives both sub-scores equal precedence while removing the necessity of a nonlinearity, such as normalization by the hyperbolic tangent. 

\begin{table}[t]
  \begin{center}
  \begin{tabular}{|c|c|}
    \hline
    WMD					& Summary Count	\\ \hline\hline
    $0 \rightarrow 1$	& $74$			\\
    $1 \rightarrow 2$	& $860$			\\
    $2 \rightarrow 3$	& $2858$		\\
    $3 \rightarrow 4$	& $2150$		\\
    $4 \rightarrow 5$	& $58$			\\
    $5+$				& $0$			\\ \hline
  \end{tabular}
  \caption{WMD among human summaries on DUC2004. For each article, every human summary was held out as the target to compare the other human summaries to resulting in 6000 comparisons.}
  \label{table:wmd}
  \end{center}
\end{table}

\subsection{Hyperparameters and Baseline}
The similarity sub-score uses a pre-trained \textit{InferSent} encoder for reproducibility, and thus needs no hyperparameter adjustments. The dissimilarity requires just the hyperparameter $\alpha$ to specify the maximum threshold of WMD and can stay at the default value of $5.0$. With the same value used to normalize the dissimilarity, VERT is straightforward to use with just this single hyperparameter. To provide a scoring reference, we test each human summary of DUC2004 on VERT using the same holdout process as done in Table \ref{table:wmd}. The average similarity sub-score is $0.7488$, the average dissimilarity sub-score is $2.7170$, and combined the average VERT score is $0.6027$.

\begin{table}[t]
  \begin{center}
  \begin{tabular}{|l|c|c|}
  \hline
  Metric  & Pearson & P-Value \\ \hline\hline
  ROUGE-1 & 0.3039  & 0.0319  \\
  ROUGE-2 & 0.2577  & 0.0708  \\
  ROUGE-L & 0.3071  & 0.0300  \\
  VERT    & 0.3681  & 0.0085  \\ \hline
  \end{tabular}
  \caption{Pearson correlation coefficient between automatic metrics and human evaluation of responsiveness.}
  \label{table:human-cor}
  \end{center}
\end{table}

\subsection{Comparison to Human Evaluation}
\label{sec:human-eval}
To evaluate the effectiveness of VERT, we calculate the correlation between VERT scores and scores given by human judges for generated sentence summaries. Using the relative dot-product attention model, 50 summaries are generated on the DUC2004 dataset and evaluated with the VERT metric by averaging the VERT scores between the four target summaries. We then conduct an experiment in which two human evaluators score the 50 generated summaries based on the DUC 2006 Responsiveness Assessment\footnote{ \url{https://duc.nist.gov/duc2007/responsiveness.assessment.instructions}}. The primary consideration of responsiveness is the amount of information in the summary that relates to the original sentence. The evaluators score the level of responsiveness on a 5-point Likert scale, with 5 being the best possible. Table \ref{table:human-cor} shows that VERT correlates with human judgment of responsiveness stronger than all three standard ROUGE metrics.


\section{Experiments}

\subsection{Experiment Setup}
The environment and evaluation of all models strictly follows the precedent set by \citet{rush2015neural}. For both training and testing, we extract sentence-summary pairs from news articles. The first sentence of each article is treated as the sentence to be summarized, while the headline of the article acts as the target summary.

\subsection{Datasets}
The training data comes from the Gigaword dataset, which is a collection of about 4 million news articles \cite{graff2003english}. It is necessary to discard certain article-headline pairs as some news articles open with a sentence that poorly relates to the headline, such as a question. Preprocessing tasks includes filtering, PTB tokenization, lower-casing, replacing digit characters with \textit{\#}, and replacing low-frequency words with \textit{UNK}. Evaluation for hyperparameter tuning is performed on the DUC2003 dataset\footnote{\url{https://duc.nist.gov/duc2003/tasks.html}}. Testing is done on the DUC2004 dataset\footnote{\url{https://duc.nist.gov/duc2004/}} where the summaries are capped at a length of 75 bytes. For both DUC2003 and DUC2004, each article has four target summaries to be compared against. For processing Gigaword, we used the same data provided by \citet{rush2015neural}, but both DUC datasets had to be preprocessed according to the tasks specified. Certain sentence-summary pairs within DUC 2004 poorly relate to each other due to the fact that the human-generated summaries used the context of the entire DUC article to decide on an adequate summary. Since this shortcoming is present across all models attempting sentence summarization on DUC, we made no effort to remove these difficult pairings from the test set.

\begin{table}[t]
\centering
\begin{tabular}{|c|r|r|r|}
\hline
Dataset		& \# Articles	& Sent Len	& Sum Len	\\ \hline\hline
Gigaword  	& 3803957    	& 31.4      & 8.3		\\
DUC2003   	& 624        	& 32.7      & 11.2  	\\
DUC2004   	& 500        	& 31.3		& 11.7  	\\ \hline
\end{tabular}
\caption{Comparison of general dataset details. Sentence and summary lengths are reported as the average word count. Gigaword has noticeably shorter target summaries than either DUC dataset. To counteract the models generating too short of summaries, we augment the beam search decoding probabilities to encourage longer summaries.}
\label{table:datasets}
\end{table}

\subsection{Base Implementation}
For the hyperparemeter specification, models used 8 attention heads and a dimension of 2048 for the dense feed forward layers. Cross entropy was used for the loss function, and optimization was performed with the Adam optimizer using a variable learning rate to encourage final convergence. Training required approximately 25 epochs. A promising feature of using an attention-based architecture is that the models used here are capable of being trained in approximately 4 hours on a single GPU, whereas recent state of the art recurrent summarization models have been mentioned to take 4 days \cite{rush2015neural}. We implemented these models using the Tensor2Tensor\footnote{\url{https://github.com/tensorflow/tensor2tensor}} library backed by TensorFlow. A strong local minimum exists when training, which closely relates to extracting the first n words of the input text up to 75 bytes. Such a trivial approach produces relatively high ROUGE scores simply due to the natural similarity between target summaries and input sentences. Diversity of attention can be encouraged by varying the learning rate and modifying the attention mechanism itself. For the decoding step, beam search is used with a beam size of 8. This results in ROUGE scores that are higher than a more simple greedy inference. Decoding to a fixed length of 75 bytes does not align easily with word-level decoding, so for the implementation we approximate the cutoff by limiting the summary sequence to 14 words.

\section{Results}

\begin{table*}[t]
  \centering
  \begin{tabular}{|l|r|r|r|c|c|c|}
    \hline
    Mechanism 			& RG-1				& RG-2			& RG-L				& VERT-S 			& VERT-D 			& VERT		\\ \hline\hline
    s-dot-prod			& 25.72				& 8.51			& 23.08				& 0.73523			& 2.76307			& 59.13		\\
    rel-s-dot-prod		& \textbf{27.05}	& \textbf{9.54}	& \textbf{24.44}	& \textbf{0.73876}	& \textbf{2.73907}	& \textbf{59.55} \\
    local				& \ \ 1.93			& 0.00			& \ \ 1.93 			& 0.02084			& 5.00000			& \ \ 1.04	\\
    local-mask			& 25.72				& 8.54			& 23.30				& 0.73361			& 2.77857			& 58.89		\\
    local-blk-mask		& 14.13				& 2.75			& 12.63				& 0.67226			& 3.18881			& 51.73		\\
    dilated				& \ \ 0.01			& \ \ 0.00		& \ \ 0.01			& 0.09509			& 3.66543			& 18.10		\\
    dilated-mask		& 19.06				& 5.23			& 17.45				& 0.68682			& 3.04922			& 53.85		\\ \hline
  \end{tabular}
  \caption{Comparison of attention mechanisms using DUC2004. RG represents ROUGE-Recall, VERT-S is the \textit{InferSent} cosine similarity sub-score, and VERT-D is the average WMD sub-score.}
  \label{table:att-compare}
\end{table*}

\begin{table*}[t]
  \begin{center}
    \begin{tabular}{|l|c|l|c|c|}
      \hline
      Model									& RG-1		& RG-2			& RG-L		& VERT	\\ \hline\hline
      TOPIARY \cite{zajic2004bbn}			& 25.12		& \ \ 6.46		& 20.12		& -		\\
      ABS \cite{rush2015neural}				& 26.55		& \ \ 7.06		& 22.05		& 58.49	\\
      RAS-LSTM \cite{chopra2016abstractive}	& 27.41		& \ \ 7.69		& 23.06		& -		\\
      MOSES+ \cite{koehn2007moses}			& 26.50		& \ \ 8.13		& 22.85		& -		\\
      RAS-Elman \cite{chopra2016abstractive}& 28.97		& \ \ 8.26		& 24.06		& -		\\
      ABS+ \cite{rush2015neural}			& 28.18		& \ \ 8.49		& 23.81		& 59.05	\\
      RA-C-LSTM \cite{zeng2016efficient} 	& 29.89		& \ \ 9.37		& 25.93		& -		\\
      words-lvt5k-1sen
      \cite{nallapati2016abstractive}		& 28.61		& \ \ 9.42		& 25.24		& -		\\
      \textbf{S-ATT-REL} (ours)				& 27.05		& \ \ 9.54		& 24.44		& 59.55	\\
      AC-ABS \cite{li2018actor}				& 32.03		& 10.99			& 27.86		& 59.67	\\ \hline
    \end{tabular}
    \caption{ROUGE-recall scores of compared models on DUC2004. Sorted by ROUGE-2 score. ABS, ABS+, and AC-ABS VERT scores were calculated using summaries provided by their respective authors.}
    \label{table:pub-compare}
  \end{center}
\end{table*}

\subsection{Attention Comparisons}
For each of the attention mechanisms described above, we performed a full scale analysis of their performance by training each model on the Gigaword dataset and evaluating on DUC2004. For each experiment, the foundational architecture was held constant. We modified both the encoder self-attention and decoder self-attention to perform as specified by the given attention mechanism. In Table \ref{table:att-compare}, the model that used scaled dot-product attention acted as the baseline (s-dot-prod). The highest performing mechanism was relative scaled dot-product attention, showing that relative positional encodings can be more insightful than absolute encodings. This demonstrates that token generation may rely more heavily on the relationships between surrounding words than relationships at a global sequential level. Local masked attention attained marginally higher ROUGE-2 and ROUGE-L scores than scaled dot-product attention. However, scaled dot-product attention scored noticeably higher with VERT, primarily due to the similarity sub-score. This suggests the scaled dot-product model is better than the local-mask model when considering the summary semantics across an entire sequence. Both local and dilated attention mechanisms repeated the same words regardless of input sentence; both masked counterparts did not have this problem.

We found a high dependence on batch size during the training process. Models would not converge when batch sizes were at or below $2000$ tokens per batch. The batch size used to train the above models was 8192 tokens. Dilated attention and dilated-mask attention models were trained at lower batch sizes due to higher memory requirements. This may have negatively effected results.

\subsection{Model Comparisons} 
We compare our best model with past work by comparing published ROUGE scores. Slight variances may be present in the reported metrics due to potential differences in data preprocessing routines. In Table \ref{table:pub-compare}, we compare our best model with that of published results. The relative dot-product self-attention model (S-ATT-REL) beats all ROUGE scores of ABS, but has a lower ROUGE-1 when ABS is tuned with an extractive routine on DUC2003 (ABS+). S-ATT-REL is comparable to but lower than certain models when it comes to ROUGE-1 scores. However, over the longer subsequence comparisons of ROUGE-2 and ROUGE-L, S-ATT-REL performs very well. This can be attributed to the ability of self-attention mechanisms to retain a strong memory over past elements of the input and decoded sequences. Only the actor-critic method (AC-ABS) beats S-ATT-REL in all tested categories.

\subsection{Qualitative Discussion}
\begin{figure}[t]
	\centering
	\includegraphics[scale=.225]{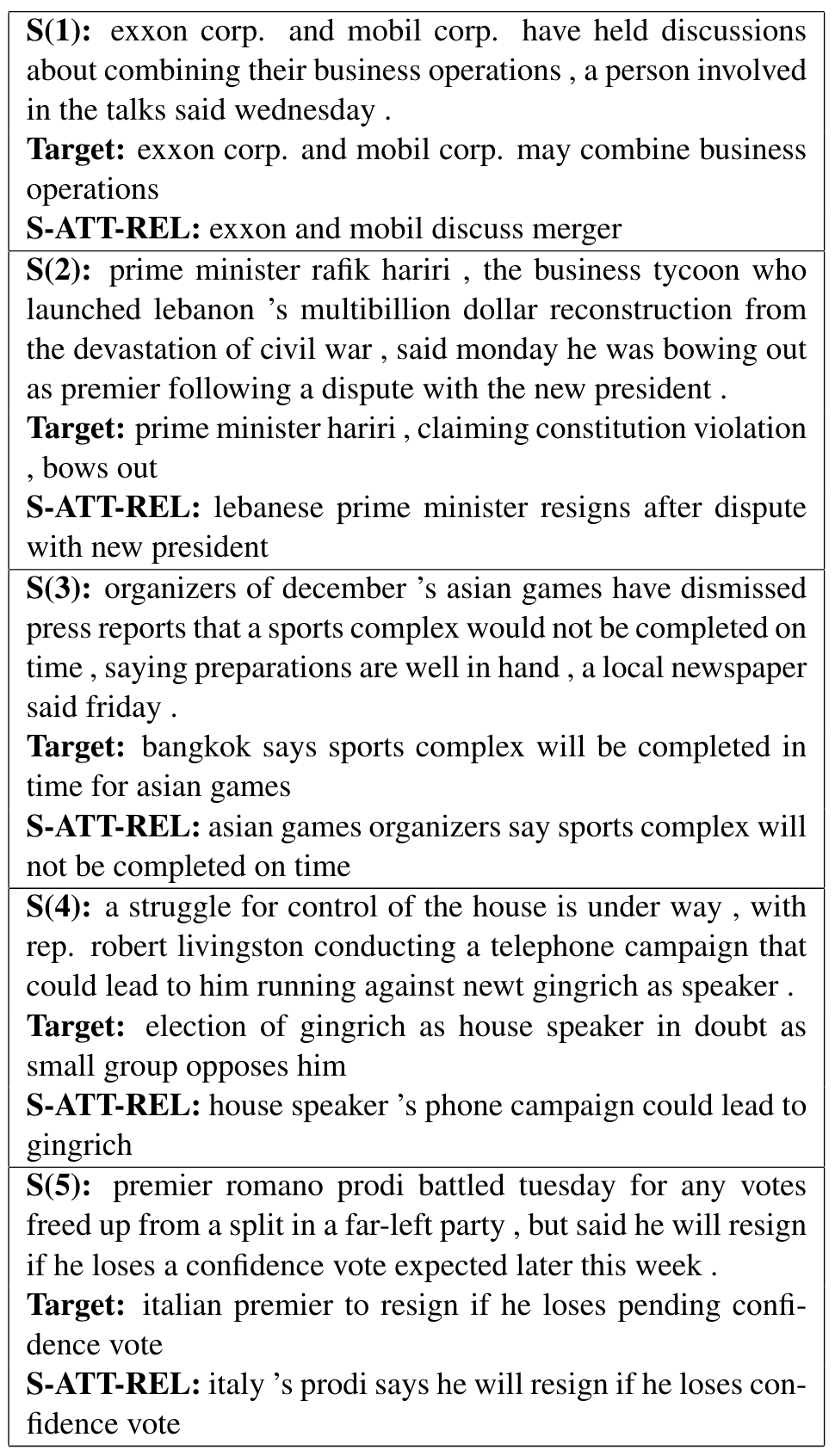}
    \caption{Examples of generated summaries by the relative dot-product self-attention model.}
    \label{fig:examples}
\end{figure}

The summaries generated by our best model are strongly abstractive, illustrated by Example \textit{S(1)} in Figure \ref{fig:examples}. Example \textit{S(2)} showcases the ability to utilize long range recall. From the appositive phrase, the model determined that Hariri was the prime minister of Lebanon and adjusted the morphology of the country for succinctness. The model also determined Hariri was resigning based on the words ``bowing out". Occasionally, attention heads are misdirected and attend to words or phrases that do not contain the primary meaning. This occurred in Example \textit{S3} with was incorrectly modified by the inclusion of ``not". The generated summaries exhibit information beyond what was directly in the input sentence; Example \textit{S5} correctly identifies Premier Romano as Italian which greatly improves the informedness of the summary. A primary strength of the self-attentive model is incorporating abstract information from all segments of the input sentence. This is suggested in the long subsequence ROUGE scores above, and seen clearly in qualitative analysis.

An assessment of linguistic quality\footnote{\url{https://duc.nist.gov/duc2007/quality-questions.txt}} was performed alongside the DUC Responsiveness Assessment. This followed the same procedure detailed in Section \ref{sec:human-eval}. Questions pertained to grammaticality, non-redundancy, referential clarity, and structure and coherence. Grammaticality scored 4.48, non-redundancy scored 4.95, referential clarity scored 4.7, and structure and coherence scored 4.53. All scores averaged between ``Good" and ``Very Good". Non-redundancy is nearly perfect, likely because the summaries are too short for redundancy to be of issue. The referential clarity scored high as well, which can be associated with the performance of the self-attention over the words already decoded.

\section{Conclusion}
The effect of modern attention mechanisms as applied to sentence summarization has been tested and analyzed. We have shown that a self-attentive encoder-decoder can perform the sentence summarization task without the use of recurrence or convolutions, which are the primary mechanisms in state of the art summarization approaches today. An inherent limitation of these existing systems is the computational cost of training associated with recurrence. The models presented can be trained on the full Gigaword dataset in just 4 hours on a single GPU. Our relative dot-product self-attention model generated the highest quality summaries among our tested models and displayed the ability of abstracting and reducing complex dependencies. We also have shown that n-gram evaluation using ROUGE metrics falls short in judging the quality of abstractive summaries. The VERT metric has been proposed as an alternative to evaluate future automatic summarization based on the premise that an abstractive summary should be judged in an abstractive manner.

\section*{Acknowledgments}
This material is based upon work supported by the National Science Foundation under Grant No. 1659788 at the University of Colorado, Colorado Springs REU site.

\bibliography{acl2018}

\begin{thebibliography}{21}
\expandafter\ifx\csname natexlab\endcsname\relax\def\natexlab#1{#1}\fi

\bibitem[{Bahdanau et~al.(2014)Bahdanau, Cho, and Bengio}]{bahdanau2014neural}
Dzmitry Bahdanau, Kyunghyun Cho, and Yoshua Bengio. 2014.
\newblock Neural machine translation by jointly learning to align and
  translate.
\newblock \emph{arXiv preprint arXiv:1409.0473}.

\bibitem[{Chopra et~al.(2016)Chopra, Auli, and Rush}]{chopra2016abstractive}
Sumit Chopra, Michael Auli, and Alexander~M Rush. 2016.
\newblock Abstractive sentence summarization with attentive recurrent neural
  networks.
\newblock In \emph{Proceedings of the 2016 Conference of the North American
  Chapter of the Association for Computational Linguistics: Human Language
  Technologies}, pages 93--98.

\bibitem[{Conneau et~al.(2017)Conneau, Kiela, Schwenk, Barrault, and
  Bordes}]{conneau2017supervised}
Alexis Conneau, Douwe Kiela, Holger Schwenk, Loic Barrault, and Antoine Bordes.
  2017.
\newblock Supervised learning of universal sentence representations from
  natural language inference data.
\newblock \emph{arXiv preprint arXiv:1705.02364}.

\bibitem[{Dong(2018)}]{dong2018survey}
Yue Dong. 2018.
\newblock A survey on neural network-based summarization methods.
\newblock \emph{arXiv preprint arXiv:1804.04589}.

\bibitem[{Gambhir and Gupta(2017)}]{gambhir2017recent}
Mahak Gambhir and Vishal Gupta. 2017.
\newblock Recent automatic text summarization techniques: a survey.
\newblock \emph{Artificial Intelligence Review}, 47(1):1--66.

\bibitem[{Ganesan(2015)}]{ganesan2015rouge}
Kavita Ganesan. 2015.
\newblock Rouge 2.0: Updated and improved measures for evaluation of
  summarization tasks.
\newblock \emph{arXiv preprint arXiv:1803.01937}.

\bibitem[{Gehring et~al.(2017)Gehring, Auli, Grangier, Yarats, and
  Dauphin}]{gehring2017convolutional}
Jonas Gehring, Michael Auli, David Grangier, Denis Yarats, and Yann~N Dauphin.
  2017.
\newblock Convolutional sequence to sequence learning.
\newblock In \emph{International Conference on Machine Learning}, pages
  1243--1252.

\bibitem[{Graff et~al.(2003)Graff, Kong, Chen, and Maeda}]{graff2003english}
David Graff, Junbo Kong, Ke~Chen, and Kazuaki Maeda. 2003.
\newblock English gigaword.
\newblock \emph{Linguistic Data Consortium, Philadelphia}, 4:1.

\bibitem[{Koehn et~al.(2007)Koehn, Hoang, Birch, Callison-Burch, Federico,
  Bertoldi, Cowan, Shen, Moran, Zens et~al.}]{koehn2007moses}
Philipp Koehn, Hieu Hoang, Alexandra Birch, Chris Callison-Burch, Marcello
  Federico, Nicola Bertoldi, Brooke Cowan, Wade Shen, Christine Moran, Richard
  Zens, et~al. 2007.
\newblock Moses: Open source toolkit for statistical machine translation.
\newblock In \emph{Proceedings of the 45th annual meeting of the ACL on
  interactive poster and demonstration sessions}, pages 177--180. Association
  for Computational Linguistics.

\bibitem[{Kusner et~al.(2015)Kusner, Sun, Kolkin, and
  Weinberger}]{kusner2015word}
Matt Kusner, Yu~Sun, Nicholas Kolkin, and Kilian Weinberger. 2015.
\newblock From word embeddings to document distances.
\newblock In \emph{International Conference on Machine Learning}, pages
  957--966.

\bibitem[{Li et~al.(2018)Li, Bing, and Lam}]{li2018actor}
Piji Li, Lidong Bing, and Wai Lam. 2018.
\newblock Actor-critic based training framework for abstractive summarization.
\newblock \emph{arXiv preprint arXiv:1803.11070}.

\bibitem[{Lin(2004)}]{lin2004rouge}
Chin-Yew Lin. 2004.
\newblock Rouge: A package for automatic evaluation of summaries.
\newblock \emph{Proceedings of the ACL Workshop: Text Summarization Branches
  Out}.

\bibitem[{Liu et~al.(2018)Liu, Saleh, Pot, Goodrich, Sepassi, Kaiser, and
  Shazeer}]{liu2018generating}
Peter~J. Liu, Mohammad Saleh, Etienne Pot, Ben Goodrich, Ryan Sepassi, Lukasz
  Kaiser, and Noam Shazeer. 2018.
\newblock \href {https://openreview.net/forum?id=Hyg0vbWC-} {Generating
  wikipedia by summarizing long sequences}.
\newblock In \emph{International Conference on Learning Representations}.

\bibitem[{Nallapati et~al.(2017)Nallapati, Zhai, and
  Zhou}]{nallapati2017summarunner}
Ramesh Nallapati, Feifei Zhai, and Bowen Zhou. 2017.
\newblock Summarunner: A recurrent neural network based sequence model for
  extractive summarization of documents.
\newblock In \emph{AAAI}, pages 3075--3081.

\bibitem[{Nallapati et~al.(2016)Nallapati, Zhou, dos Santos,
  G{\"{u}}l{\c{c}}ehre, and Xiang}]{nallapati2016abstractive}
Ramesh Nallapati, Bowen Zhou, C{\'{\i}}cero~Nogueira dos Santos, {\c{C}}aglar
  G{\"{u}}l{\c{c}}ehre, and Bing Xiang. 2016.
\newblock \href {http://aclweb.org/anthology/K/K16/K16-1028.pdf} {Abstractive
  text summarization using sequence-to-sequence rnns and beyond}.
\newblock In \emph{Proceedings of the 20th {SIGNLL} Conference on Computational
  Natural Language Learning, CoNLL 2016, Berlin, Germany, August 11-12, 2016},
  pages 280--290.

\bibitem[{Paulus et~al.(2017)Paulus, Xiong, and Socher}]{paulus2017deep}
Romain Paulus, Caiming Xiong, and Richard Socher. 2017.
\newblock A deep reinforced model for abstractive summarization.
\newblock \emph{arXiv preprint arXiv:1705.04304}.

\bibitem[{Rush et~al.(2015)Rush, Chopra, and Weston}]{rush2015neural}
Alexander~M Rush, Sumit Chopra, and Jason Weston. 2015.
\newblock A neural attention model for abstractive sentence summarization.
\newblock In \emph{Proceedings of the 2015 Conference on Empirical Methods in
  Natural Language Processing}, pages 379--389.

\bibitem[{See et~al.(2017)See, Liu, and Manning}]{see2017get}
Abigail See, Peter~J. Liu, and Christopher~D. Manning. 2017.
\newblock \href {https://doi.org/10.18653/v1/P17-1099} {Get to the point:
  Summarization with pointer-generator networks}.
\newblock In \emph{Proceedings of the 55th Annual Meeting of the Association
  for Computational Linguistics (Volume 1: Long Papers)}, pages 1073--1083.
  Association for Computational Linguistics.

\bibitem[{Vaswani et~al.(2017)Vaswani, Shazeer, Parmar, Uszkoreit, Jones,
  Gomez, Kaiser, and Polosukhin}]{vaswani2017attention}
Ashish Vaswani, Noam Shazeer, Niki Parmar, Jakob Uszkoreit, Llion Jones,
  Aidan~N Gomez, {\L}ukasz Kaiser, and Illia Polosukhin. 2017.
\newblock Attention is all you need.
\newblock In \emph{Advances in Neural Information Processing Systems}, pages
  6000--6010.

\bibitem[{Zajic et~al.(2004)Zajic, Dorr, and Schwartz}]{zajic2004bbn}
David Zajic, Bonnie Dorr, and Richard Schwartz. 2004.
\newblock Bbn/umd at duc-2004: Topiary.
\newblock In \emph{Proceedings of the HLT-NAACL 2004 Document Understanding
  Workshop, Boston}, pages 112--119.

\bibitem[{Zeng et~al.(2016)Zeng, Luo, Fidler, and Urtasun}]{zeng2016efficient}
Wenyuan Zeng, Wenjie Luo, Sanja Fidler, and Raquel Urtasun. 2016.
\newblock Efficient summarization with read-again and copy mechanism.
\newblock \emph{arXiv preprint arXiv:1611.03382}.

\end{thebibliography}
\bibliographystyle{acl_natbib}

\end{document}